# Energy Efficient Personalized Hand-Gesture Recognition with Neuromorphic Computing

Muhammad Aitsam
Sheffield Hallam University, UK
United Kingdom
m.aitsam@shu.ac.uk

Alessandro di Nuovo
Sheffield Hallam University
United Kingdom
a.dinuovo@shu.ac.uk

## ABSTRACT

Hand gestures are a form of non-verbal communication that is used in social interaction and it is therefore required for more natural human-robot interaction. Neuromorphic (brain-inspired) computing offers a low-power solution for Spiking neural networks (SNNs) that can be used for the classification and recognition of gestures. This article introduces the preliminary results of a novel methodology for training spiking convolutional neural networks for hand-gesture recognition so that a humanoid robot with integrated neuromorphic hardware will be able to personalise the interaction with a user according to the shown hand gesture. It also describes other approaches that could improve the overall performance of the model.

## 1 INTRODUCTION

Hand gesture recognition is an active field of research driven by the desire to provide a more efficient, intuitive, and better personlized human-robot interaction (HRI) [17]. To address this challenge, various sensor modes have been employed, including radar, electromyography systems (EMG), cameras, dynamic vision systems (DVS), etc. There has also been a variety of learning techniques proposed to tackle hand gesture recognition, including convolutional neural networks (CNNs) [1], long short-term memories (LSTMs) [8], and spiking neural networks [21]. This article focuses on the spiking neural networks (SNNs) technique to train a personalized hand gesture recognition model for better HRI. In hand gesture recognition for human-robot interaction, personalization refers to the process of adapting the recognition system to a specific user's hand gestures. This is achieved by fine-tuning the recognition model with data collected from that user. Personalization can improve the accuracy of the recognition system for that user by taking into account variations in hand shape, size, and movement that may be unique to that individual. Personalization can enable the system to better understand and respond to a user's unique hand gestures, making the interaction more natural and intuitive. It can also improve the robustness of the system to variations in lighting, background, and hand pose, making it more suitable for use in real-world environments [12].

In the last few decades, several scientific and commercial applications have benefited from the increased power of computers and sophisticated sensing systems. Although robots nowadays are somewhat capable of performing human-like tasks, they lack precise motor control, fast reaction times, and adaptation to changing conditions [2]. Moreover, current machine learning models also lack scalability. The difference between current technology and human brain can be exemplified by the fact that a hypothetical computer running a human-scale brain simulation requires around 12 Gigawatt of power, contrary to human brain which works only with 20 Watt [6]. To sense and act like living beings, the clock frequencies of the system must be increased to deal with continuous input from the real world. However, very high frequencies make current hardware inefficient for large-scale applications.

The biological behaviour of the human brain, which uses extremely little energy and reacts very quickly, has been studied extensively in this area of concentration in recent years, and numerous new technologies and methods are being created in an effort to replicate it. Neuromorphic computing is one of this technology [19]. An interdisciplinary research paradigm called neuromorphic computing or brain-inspired computing examines massively parallel processing devices that enable spike-driven communication in natural neural computations. The main benefits of neuromorphic computing over conventional methods are its energy efficiency, speed of execution, and robustness against local failures. Emerging hardware and software expertise in the fields of neuroscience and electronics has made it possible to construct biologically inspired robots by modelling cognitive and interactive capabilities using spiking neural networks (SNNs), which are inspired by this event-driven mode of information processing [15].

Both neuromorphic computing and robotics are heavily reliant on human-machine interaction. Utilising neuromorphic technologies in robotics is a promising approach to creating robots that can seamlessly integrate into society. In this work, we are training a multi-layer spiking neural network model for personalized hand-gesture recognition to improve human-robot interaction.

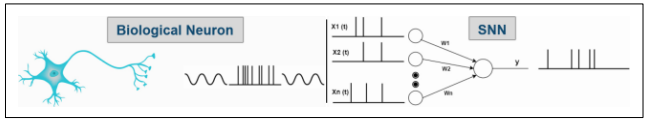

Figure 1: The simplified comparison between biological neuron and neuron in a spiking neural network.

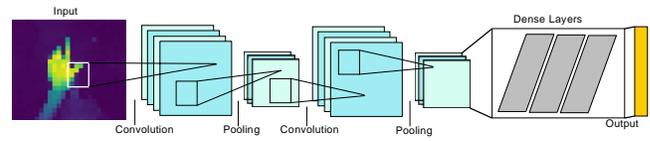

Figure 2: The topology of CNN for training hand gesture dataset. The model consists of an input layer followed by intermediate hidden layers and a fully connected layer at the end.

In the next section, we briefly discuss spiking neural networks (SNNs) and how learning happens in this paradigm. Section 3 gives a brief description of the convolutional neural network (CNN) and what lessons can be learned from it to train SNN models. Section 4 is about the methodology we used for our experiments. The results are discussed in section 5. Section 6 discusses future work. Section 7 gives our conclusions.

## 2 SPIKING NEURAL NETWORKS (SNNS)

Spiking Neural Networks (SNNs) are often regarded as the third generation of neural networks that can be highly power-efficient and have capabilities to perform cognitive tasks. SNN imitates the action of neurons in the brain. In brain, the neurons are excitable and produce spikes, also known as an action potential. These spikes which are the basic currency of the brain, allow a neuron to process the information and communicate with other neurons. SNNs also operate on spikes which are discrete events taking place at a specific time. Reduced computational complexity, temporal plasticity, and ease of use of neural interfaces [9] are some of the major advantages of SNNs. Figure 1 shows the similarity between SNN and biological neurons. However, the key concepts of SNN operations do not allow the use of classical learning techniques and methods that are appropriate for a CNN but there are still several methods to train SNNs [11], [4], [3], [18].

The most well-known technique for unsupervised learning in SNNs is spike-timing-dependent plasticity (STDP) [10]. The difference in firing rates between pre- and post-synaptic neurons serves as the basis for the synaptic weight in STDP. Long-term potentiation (LTP) of the synapses occurs when a pre-synaptic spike occurs before the post-synaptic action potential, whereas LTD occurs when a pre-synaptic spike occurs after the post-synaptic spike. The STDP function also referred to as the learning window, is the change in the synapse displayed as a function of the relative timing of pre-and postsynaptic events. In this work, we are using this learning technique for hand-gesture recognition during human-robot interaction (HRI).

## 3 CONVOLUTIONAL NEURAL NETWORKS (CNNS)

CNNs are distinguished from other neural networks because of their superior performance with speech and image inputs [7]. It has three main types of layers, convolutional layer, pooling layer and fully-connected layer. For the recognition of high-dimensional input, patterns necessitate a multi-layer network to learn from input stimuli [14]. In this work, we use multi-layer CNN that consist of two convolutional layers, two pooling layer and three dense layers. The convolutional layers' weight kernels encode the feature representations at various levels of hierarchy. Due to convolution, which by nature renders the network invariant to translation (shift) in the object location, the updated convolutional kernels can appropriately detect the spatially associated local characteristics in the input patterns. Next, the pooling layers offer the following advantages. First, it reduces the dimension of the convolutional feature maps while adding a tiny bit of additional network invariance to input transformation. Additionally, the pooling procedure increases the effective size of convolutional kernels in the following layer by virtue of downscaling the feature maps. This facilitates the effective learning of hierarchical representations at various levels of abstraction by succeeding convolutional layers. Lastly, The fully-connected layer functions as a classifier to efficiently include the feature composition from the previous layers. The complete CNN topology is illustrated in Figure 2.

## 4 METHODOLOGY

This section covers the methodology we use to conduct experiments. Besides this, we also discuss the neuromorphic hardware used in the process. The open-source hand gesture dataset is selected to train CNN. The dataset is composed of 10 different hand gestures that were performed by 10 different subjects (5 men and 5 women) [13]. First, we trained CNN model for our selected dataset. After passing through alternating convolution layers, pooling layers and finally with a fully connected layer, our trained model achieved 99.8 percent accuracy. The final CNN model is saved. In the next step, we convert our trained CNN model to the SNN model. For this purpose, we use the SNN toolbox (SNN-TB) [22] which is by far one of the best frameworks to transform rate-based artificial neural networks into spiking neural networks. SNN-TB accepts input models from different deep learning libraries like PyTorch, TensorFlow, Keras, etc., and also provides backend interfaces like PyNN, Nest, Neuron, etc. These accepted models can also be deployed to neuromorphic chips like SpiNNaker [16] and Loihi [5].

In our experiment, we use 48 chip SpiNNaker board. Based on billions of simple computing elements communicating via unreliable spikes, SpiNNaker is a new massively parallel computer architecture inspired by the biology of the brain. Before the conversion, we did the pre-processing of our dataset to make it usable for SNN training. SNN-TB only accepts data in NumPy zip (.npz) format, so we transformed the data and did the train-test split. In the next step, we created a config file for SNN-TB. Here we set batch size as 8, the total duration to simulate each sample is 200ms, and we run the test 20 times. We set the ANN evaluation as True and also provided the Poisson input with a rate of 1000. Figure 3 shows the configuration file parameters and block diagram of the system.

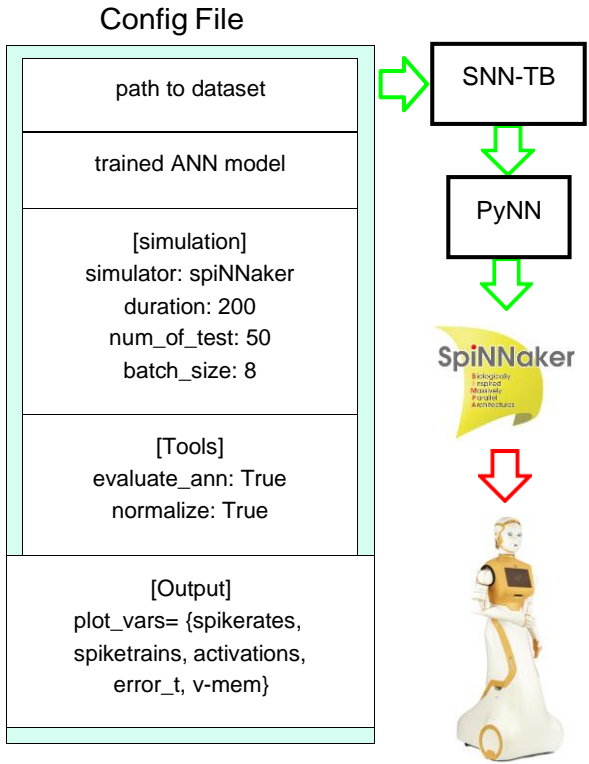

**Figure 3: The configuration file parameters and block diagram of the whole system. At this stage, we are not conducting experiments with robots.**

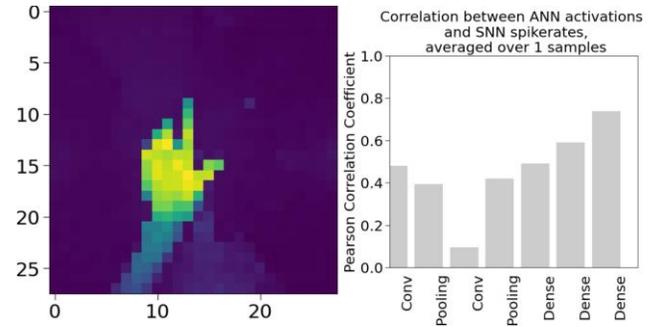

**Figure 4: One of the input images from pre-processed dataset and spiketrain of final layer. The correlation coefficient of each layer in ANN with respect to SNN spikerate. Here we can see that the final dense layer has the highest correlation coefficient.**

## 5 RESULTS AND DISCUSSION

The goal of the experiments is to convert the unsupervised trained CNN model to SNN model for hand gesture recognition which can be used in human-robot interaction. With our current approach, after running the experiments on the SpiNNaker board through PyNN, we find out that the final accuracy of our model is not up to the mark. Figure 4 shows the input image (class 10) and the correlation between ANN activation and SNN spike rates. Here we can see that the final layer has the highest correlation and the second convolution layer shows the lowest correlation coefficient. Although the overall performance of the model with this approach is not that great, SpiNNaker consumes far less energy to execute the model as compared to conventional CPUs and GPUs [20].

We are still investigating the reasons behind this low accuracy. One of the reasons could be the parameters we selected during the configuration of SNN-TB. There is no exact formula available to choose the optimal parameters. First, we are using a trial-and-error approach to fine-tune the parameters. Secondly, we are training CNN for other hand gesture datasets to compare the results. Lastly, we are also looking for other approaches to train spiking convolutional neural networks for hand-gesture recognition. These approaches are discussed in the next section.

## 6 FUTURE WORK

In this preliminary study, we used the ANN-to-SNN conversion approach for the hand-gesture recognition task. In the next phase of this study, we are going to fine-tune the parameters to increase the overall accuracy of the converted SNN model. Besides this, we are planning to use two more approaches for this task.

The first approach is to get the synaptic connections from the ANN activation in each layer and use this connection to train our SNN model. It has been seen that conversion frameworks (SNN-TB) give better results for software simulators (Nest and Neuron) as compared to hardware simulators (SpiNNaker). SNN-TB generates connection files for every layer during conversion. Each file contains four columns (index of pre-neuron, index of post-neuron, weight, and delay). We are going to use Neuron or Nest simulator to get these connections and then use them with our SpiNNaker board. This approach is not been tested before so only the results will tell its effectiveness.

The second approach is to train the SNN model from scratch. Here we have to use existing learning techniques like STDP for training. We plan to use the lessons from CNN to train the network. Instead of going directly from the input to the excitatory layer and inhibitory layer to get the output, we will introduce hidden layers similar to CNN. The output after passing the input through the first hidden layer will be the input for the next layer and at the end with the fully connected layer, the network will be able to recognize the hand gesture.

The goal of human-robot collaboration (HRI) is to combine the advantages of robots with the flexibility and cognitive skills of humans to enable them to work together in a shared environment. In the final part of project, we will use a humanoid robot (ARI). After training the model, we expect a robot to get real-time input from the user standing in front of it, showing hand-gesture. The robot should be able to identify the hand gesture and respond accordingly to improve personalized human-robot interaction. Some of the expected challenges are connecting the spiking camera to the robot and integrating the humanoid robot with neuromorphic hardware.

## 7 CONCLUSION

In recent years, neuromorphic computing with spiking neural networks has demonstrated remarkable results in terms of computational speed, energy efficiency, and human-like intelligence. Many robotics-related cognitive applications are also being developed in this domain. In this article, we have shown a spiking convolutional neural network approach where the pre-trained CNN model is converted to an SNN model by using the SNN toolbox for personalized hand-gesture recognition tasks. The 48-chip SpiNNaker board is used for the SNN model. We discussed other approaches to train the SNN model efficiently for our desired task and how our final SNN model, when integrated with a humanoid robot, can improve human-robot interaction.

## ETHICAL STATEMENT

The work described here did not involve human participants. Data used has been retrieved from publicly available datasets.

## ACKNOWLEDGMENT

This work is funded by Marie Sklodowska-Curie Action Horizon 2020 (Grant agreement No. 955778) for the project 'Personalized Robotics as Service Oriented Applications (PERSEO)' and by the UK EPSRC with the grant EP/W000741/1 (EMERGENCE).